\pdfoutput=1
\documentclass[11pt]{article}
\usepackage[final]{acl}
\usepackage{times}
\usepackage{latexsym}
\usepackage[T1]{fontenc}
\usepackage[utf8]{inputenc}
\usepackage{microtype}
\usepackage{inconsolata}
\usepackage{graphicx}
\usepackage{afterpage}
\usepackage{float}
\usepackage{multirow} 
\usepackage{booktabs}
\usepackage{colortbl}
\usepackage{xcolor}
\usepackage{amsmath}
\usepackage{amsfonts}  
\usepackage{amssymb}
\usepackage[T1]{fontenc}
\usepackage{graphicx}
\usepackage{algorithmic}
\usepackage{xcolor}
\usepackage{multirow}
\usepackage{booktabs}
\usepackage{color}
\usepackage{algorithm2e}
\usepackage{amsmath}
\usepackage{latexsym}
\usepackage[utf8]{inputenc}
\usepackage{microtype}
\usepackage{inconsolata}
\usepackage{afterpage}
\usepackage{float}
\usepackage{multirow} 
\usepackage{booktabs}
\usepackage{colortbl}
\usepackage{xcolor}
\usepackage{amsfonts}  
\usepackage{amssymb}
\usepackage{pifont}
\usepackage{enumitem}
\definecolor{lightgray}{rgb}{0.95, 0.95, 0.95}
\usepackage{placeins}
\newcommand{\method}{ToM} 
\newcommand{\deepseeklogo}{%
    \includegraphics[height=0.9em]{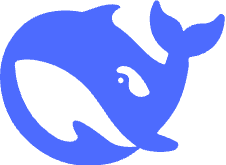}%
}

\newcommand{\qwenlogo}{%
    \includegraphics[height=1.12em]{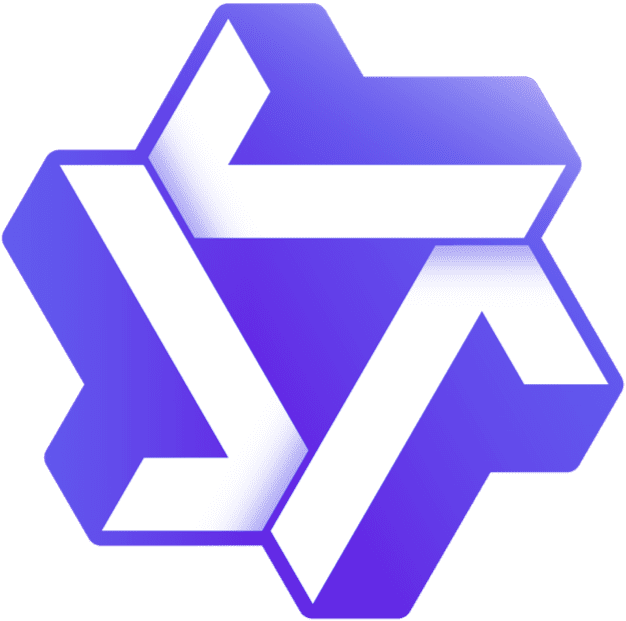}%
}

\newcommand{\gptlogo}{%
    \includegraphics[height=1.00em]{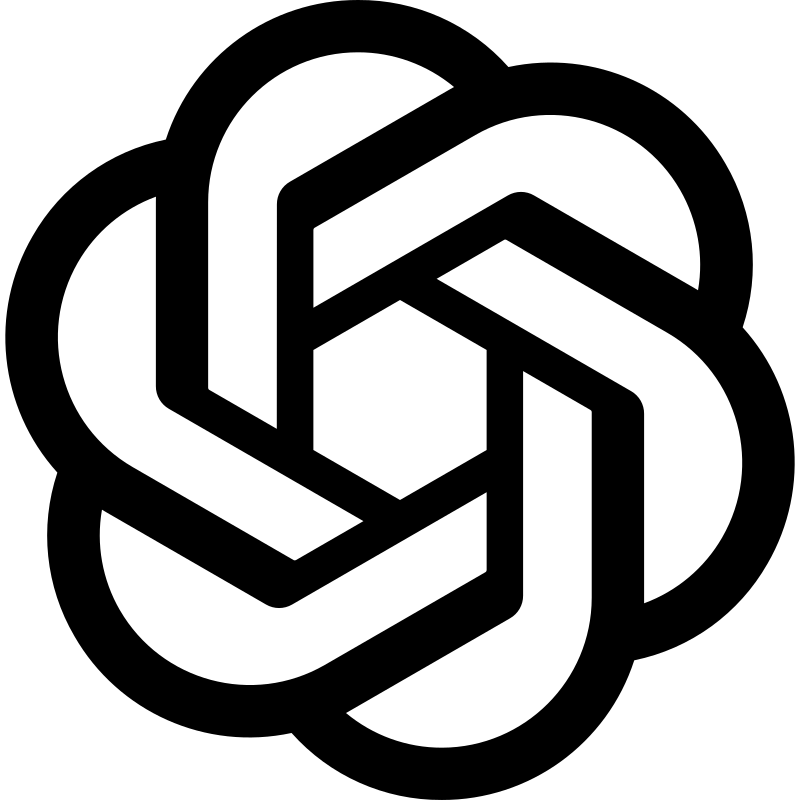}%
}

\newcommand{\llamalogo}{%
    \includegraphics[height=1.2em]{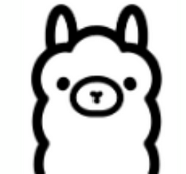}%
}

\newcommand{\deepseeklogowithtext}{%
    \raisebox{-0.35ex}{\deepseeklogo}\,Deepseek-V3%
}

\newcommand{\deepseeklogowithMoD}{%
    \raisebox{-0.35ex}{\deepseeklogo}\,+ ToM%
}

\newcommand{\deepseekRlogowithMoD}{%
    \raisebox{-0.35ex}{\deepseeklogo}\,R1 + ToM
}

\newcommand{\deepseekRlogowithtext}{%
    \raisebox{-0.35ex}{\deepseeklogo}\,Deepseek-R1 
}

\newcommand{\deepseekRlogowithLA}{%
    \raisebox{-0.35ex}{\deepseeklogo}\,R1 + LongAgent
}

\newcommand{\deepseeklogowithLA}{%
    \raisebox{-0.35ex}{\deepseeklogo}\,+ LongAgent%
}

\newcommand{\qwenlogowithtext}{%
    \raisebox{-0.35ex}{\qwenlogo}\,Qwen2.5-72B-Instruct%
}

\newcommand{\gptlogowithtext}{%
    \raisebox{-0.35ex}{\gptlogo}\, GPT-4o%
}

\newcommand{\gptlogowithMoD}{%
    \raisebox{-0.35ex}{\gptlogo}\, + ToM%
}

\newcommand{\gptlogowithLA}{%
    \raisebox{-0.35ex}{\gptlogo}\, + LongAgent%
}

\newcommand{\llamalogowithtext}{%
    \raisebox{-0.35ex}{\llamalogo}\,Llama3-70B-Instruct%
}

\newcommand{\llamalogowithLA}{%
    \raisebox{-0.35ex}{\llamalogo}\, + LongAgent%
}

\newcommand{\qwenlogowithLA}{%
    \raisebox{-0.35ex}{\qwenlogo}\, + LongAgent%
}

\newcommand{\llamalogowithToM}{%
    \raisebox{-0.35ex}{\llamalogo}\, + ToM%
}

\newcommand{\qwenlogowithToM}{%
    \raisebox{-0.35ex}{\qwenlogo}\, + ToM%
}

\title{ToM: Leveraging Tree-oriented MapReduce for Long-Context Reasoning in Large Language Models}

\author{
 \textbf{Jiani Guo\textsuperscript{1}\footnotemark[1]},
 \textbf{Zuchao Li\textsuperscript{2}\footnotemark[2]},
 \textbf{Jie Wu\textsuperscript{3}\footnotemark[1]},
 \textbf{Qianren Wang\textsuperscript{4}},
 \\
 \textbf{Yun Li\textsuperscript{5}},
 \textbf{Lefei Zhang\textsuperscript{1}},
 \textbf{Hai Zhao\textsuperscript{6}},
 \textbf{Yujiu Yang\textsuperscript{3}}
\\ 
 \textsuperscript{1}School of Computer Science, Wuhan University, Wuhan, China \\
 \textsuperscript{2}School of Artificial Intelligence, Wuhan University, Wuhan, China \\
 \textsuperscript{3}Tsinghua University;
 \textsuperscript{4}Shanghai  Huawei Technologies, China;
 \textsuperscript{5}Cognitive AI Lab \\
 \textsuperscript{6}School of Computer Science, Shanghai Jiao Tong University, Shanghai, China \\
 \texttt{\{guojiani, zcli-charlie, zhanglefei\}@whu.edu.cn} \\
\\
}

\begin{document}
\maketitle
\begin{abstract}
 Large Language Models (LLMs), constrained by limited context windows, often face significant performance degradation when reasoning over long contexts.
 To address this, Retrieval-Augmented Generation (RAG) retrieves and reasons over chunks but frequently sacrifices logical coherence due to its reliance on similarity-based rankings.
 Similarly, divide-and-conquer frameworks (DCF) split documents into small chunks for independent reasoning and aggregation.
 While effective for local reasoning, DCF struggles to capture long-range dependencies and risks inducing conflicts by processing chunks in isolation. To overcome these limitations, we propose ToM, a novel Tree-oriented MapReduce framework for long-context reasoning.
    ToM leverages the inherent hierarchical structure of long documents (e.g., main headings and subheadings) by constructing a DocTree through hierarchical semantic parsing and performing bottom-up aggregation. Using a Tree MapReduce approach, ToM enables recursive reasoning: in the \textbf{Map} step, rationales are generated at child nodes; in the \textbf{Reduce} step, these rationales are aggregated across sibling nodes to resolve conflicts or reach consensus at parent nodes. Experimental results on 70B+ LLMs show that ToM significantly outperforms existing divide-and-conquer frameworks and retrieval-augmented generation methods, achieving better logical coherence and long-context reasoning. Our code is available at \href{https://github.com/gjn12-31/ToM}{https://github.com/gjn12-31/ToM}.
    
\end{abstract}

\renewcommand{\thefootnote}{\fnsymbol{footnote}}
\footnotetext[1]{Equal contribution.}
\footnotetext[2]{Corresponding Author. }

\section{Introduction}
Large Language Models (LLMs) with limited context windows (e.g., 8k, 32k) struggle with reasoning over long contexts. As context length increases, the performance declines due to difficulties in processing information far from the text's beginning or end~\cite{liu-etal-2024-lost,DBLP:conf/acl/HePDSLQLWZZ24}.
To address these limitations, two mainstream training-free approaches have been introduced: Retrieval-Augmented Generation (RAG)~\cite{DBLP:conf/emnlp/Li00MB24,DBLP:journals/corr/abs-2410-05779} and Divide-and-Conquer Frameworks (DCF)~\cite{zhao-etal-2024-longagent,zhou2024llmtimesmapreducesimplifiedlongsequenceprocessing,DBLP:journals/corr/abs-2406-02818}.

RAG addresses the challenge of long-context reasoning by using a retriever to identify the most relevant chunks from a long document, focusing only on the most pertinent information for reasoning.
However, RAG relies on similarity rankings and often neglects the logical coherence between retrieved chunks.
In contrast, DCF processes long contexts by splitting them into smaller chunks, reasoning over each independently, and synthesizing local insights into a global understanding.
While effective to some extent, DCF treats chunks in isolation and overlooks the relationships between non-adjacent or long-range segments, leading to conflicts and incomplete understanding due to its limited local perspective.
Inspired by the idea that human thinking is inherently complex and multidimensional, extending beyond what can be captured by flat reasoning or independent reasoning processes~\cite{Barsalou_1999}.
Reflecting this complexity, long contexts naturally exhibit a hierarchy, such as main headings and subheadings, well suited for reasoning with a tree-based representation.

\begin{figure*}
    \centering
    \includegraphics[width=1.0\linewidth]{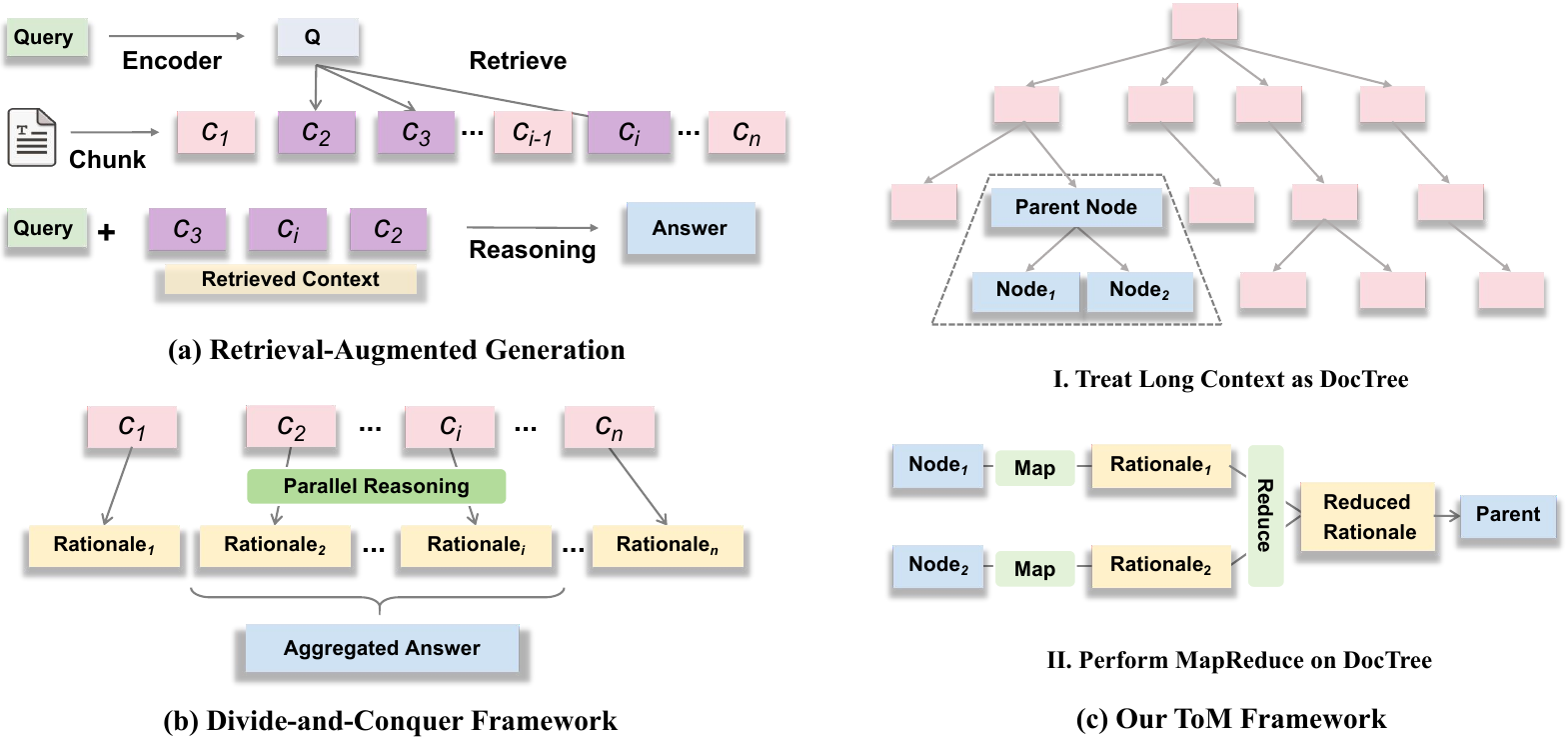}
    \caption{Comparison between ToM and existing approaches:
    LLMs enhanced with RAG (a) rely on sequential reasoning over retrieved chunks, while current Divide-and-Conquer frameworks (b) process chunks in isolation. In contrast, ToM (c) leverages the hierarchical structure of DocTree for tree-based reasoning, mitigating conflicts and preserving contextual coherence through recursive MapReduce reasoning.}
    \label{fig:motivation}
\end{figure*}

We introduce ToM, a novel framework for tree-based long-context reasoning.
Leveraging a tree-structured MapReduce approach, ToM performs recursive reasoning over documents to enhance long-context understanding. It consists of two key components: 1). \textbf{DocTree Construction}: ToM first applies Hierarchical Semantic Parsing to convert each chunk into a structured subtree, then combines these subtrees into a hierarchical DocTree through Bottom-up Aggregation. 2). \textbf{Recursive Reasoning via MapReduce}: ToM performs recursive reasoning on the DocTree in a MapReduce fashion, enabling systematic aggregation of rationales across the hierarchy.
In the \textit{Map} step, ToM generates rationales from child nodes; in the \textit{Reduce} step, these rationales are aggregated across sibling nodes to resolve conflicts or reach consensus at the parent node.
As shown in Figure~\ref{fig:motivation}, ToM enables focal reasoning across the hierarchy, enhancing information utilization beyond the flat reasoning employed by RAG. Compared to DCF, ToM considers sibling and parent-child relationships between chunks and allows long-range chunks to connect to the same parent node through bottom-up reasoning, thereby improving fact aggregation and reducing conflicts.

Experimental results on 70B+ LLMs show that ToM achieves significant performance gains over current divide-and-conquer methods and LLMs enhanced with retrieval-augmented generation.
Further analysis shows that ToM balances efficiency and effectiveness by combining DocTree compression with embedding techniques.

Our contributions are shown as follows:
\begin{enumerate}
 \item \textbf{DocTree Representation.} 
 We propose DocTree, a hierarchical representation for organizing long documents. This structure leverages Hierarchical Semantic Parsing and Bottom-up Aggregation to transform chunked text into well-structured trees for bottom-up reasoning.
  \item \textbf{ToM Framework.} We introduce ToM, a tree-oriented MapReduce framework for long-context reasoning.
By leveraging parent-child and sibling relationships, ToM enables structured reasoning, resolves conflicts more effectively, and improves information utilization.
\item \textbf{Experimental Validation.} Experiments on 70B+ LLMs demonstrate that ToM outperforms existing divide-and-conquer methods and RAG-enhanced LLMs, achieving significant performance gains in long-context reasoning. Comprehensive ablation studies, complexity analysis, and case studies collectively highlight ToM's effectiveness.
\end{enumerate}

\section{Related Work}
\noindent \textbf{Long-context Reasoning}~\cite{DBLP:conf/acl/LiWZZ24,DBLP:conf/acl/LiLL024,hu-etal-2025-longrecipe} has become essential, as LLMs with limited context windows often suffer performance degradation in long-input scenarios\cite{liu-etal-2024-lost}.
Training-based methods, such as LongLoRA~\cite{chen2024longlora}, extend context windows via fine-tuning using position encoding refinement, sparse attention, or learnable embeddings. While effective, they demand large datasets support and high computational cost.
Training-free methods, like InfLLM~\cite{xiao2024infllm}, retrieve token-relevant content from memory to optimize attention. Divide-and-conquer strategies such as LongAgent~\cite{zhao-etal-2024-longagent}, LLM$\times$MapReduce~\cite{zhou2024llmtimesmapreducesimplifiedlongsequenceprocessing}, and Chain-of-Agents~\cite{DBLP:journals/corr/abs-2406-02818} split long contexts into chunks, process them locally, and aggregate the results.

\noindent \textbf{MapReduce}~\cite{DBLP:journals/cacm/DeanG08,langchain} simplifies large-scale data processing by dividing tasks into two phases: the Map step processes data into key-value pairs, and Reduce aggregates the results.
LongChain first introduces MapReduce for multi-document scenarios, applying the Map step to each document and combining outputs into a single result via Reduce.
Methods like LLM$\times$MapReduce\cite{zhou2024llmtimesmapreducesimplifiedlongsequenceprocessing} and XL$^3$M~\cite{DBLP:journals/corr/abs-2405-17755} extend this to long documents by splitting them into sub-contexts, selecting relevant segments, and combining them chronologically.
The major challenge for MapReduce is that chunks are processed independently, which may break essential long-range dependencies and interconnections between them.

\noindent \textbf{Retrieval Augmented Generation} (RAG)~ \cite{DBLP:conf/nips/LewisPPPKGKLYR020,jeong-etal-2024-adaptive,DBLP:conf/acl/WangTOXS24,wang-etal-2024-searching,DBLP:conf/iclr/Sun0TYZ23,edge2024localglobalgraphrag,DBLP:journals/corr/abs-2410-05779,zeng2025FSDrive,DBLP:conf/iclr/AsaiWWSH24, xia-etal-2024-rule,liu2024rag,liu2025qfft,zhang2025rearank,zhang2024exploring} enhances response quality by combining retrieval and in-context learning.
% Self-RAG \cite{DBLP:conf/iclr/AsaiWWSH24} adds self-reflection to refine retrieval choices. 
% GraphRAG \cite{edge2024localglobalgraphrag} and LightRAG \cite{DBLP:journals/corr/abs-2410-05779} incorporate graph knowledge to handle complex relationships. 
Inspired by the idea of recursive summarization in Raptor~\cite{DBLP:conf/iclr/SarthiATKGM24}, we go further by capturing hierarchical relationships within each chunk and applying bottom-up aggregation, aiming to represent the long document as a structured DocTree. While Raptor uses summary nodes for retrieval-augmented generation, we employ detailed, recursive bottom-up reasoning with a MapReduce approach, fully utilizing the constructed tree.

\begin{figure*}
    \centering
    \includegraphics[width=1.\linewidth]{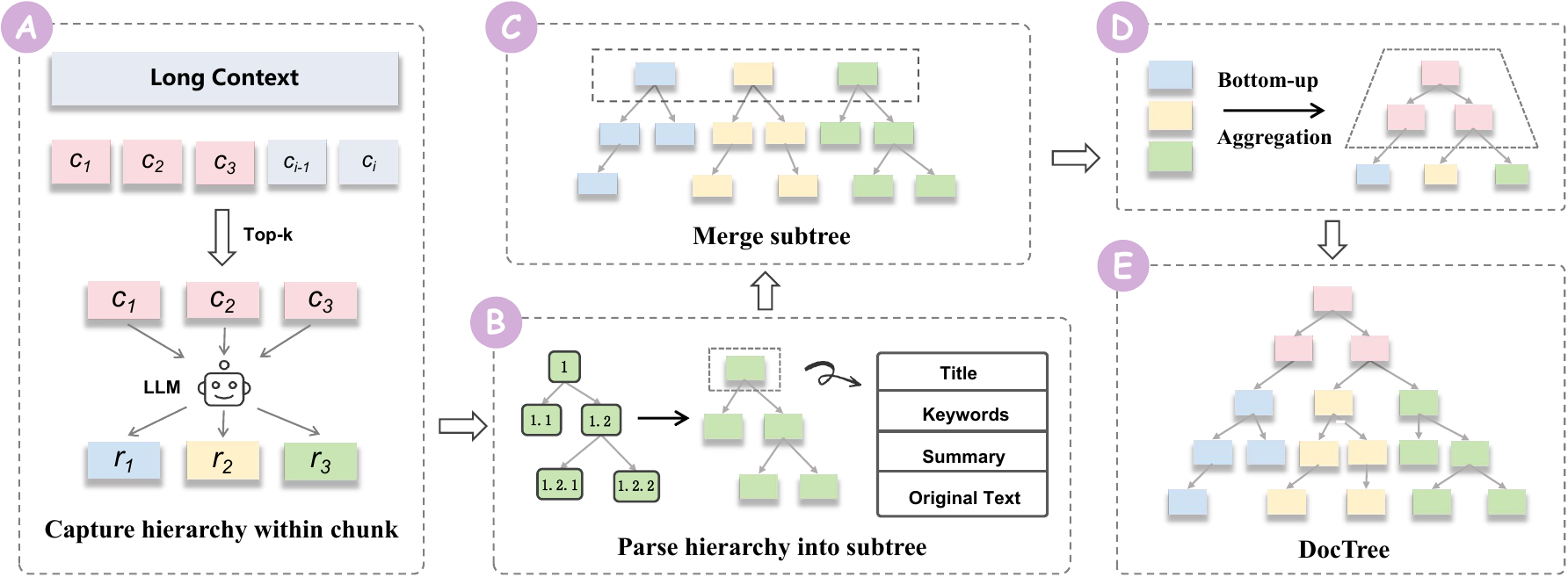}
    \caption{Illustration for DocTree Construction: The process begins with preparation for compression, where a retriever selects Top-k chunks as the foundation for the tree. Each chunk is then processed (A) using the Hierarchical Semantic Parser to capture its internal hierarchy. Next, (B) the hierarchical structure of each chunk is parsed into a subtree, with nodes capturing structured information.
    These subtrees are merged (C) by placing root nodes at the same level. Following this, (D) a Bottom-up Aggregation integrates information across levels. Finally, the complete DocTree is constructed (E) by combining low-level subtrees with higher-level summaries, ready for reasoning.}
    \label{fig:Tree Construct}
\end{figure*}
\section{ToM Framework}
\subsection{Overview.}
For effective tree-based reasoning, ToM employs the following two steps:
1) \textbf{Representing long contexts as DocTree (Section~\ref{sec:DocTree})}. ToM begins with \textit{Hierarchical Semantic Parsing} to structure each chunk into a subtree, where lower levels correspond to subheadings and higher levels summarize main headings. It then applies \textit{Bottom-up Aggregation} to merge subtrees into a higher-level hierarchy. 2) \textbf{Recursive MapReduce Reasoning (Section~\ref{sec:Mapreduce})}. ToM performs recursive reasoning on the DocTree using a MapReduce-style process. This involves iteratively applying the Map and Reduce steps across the hierarchy: the \textit{Map} step transforms information from child nodes into rationales that capture key supporting facts, while the \textit{Reduce} step aggregates sibling rationales to resolve conflicts and reach consensus.
By applying recursive MapReduce to the DocTree, ToM enables focused fact aggregation and conflict resolution, thereby facilitating effective long-context reasoning.

\subsection{Representing Long Contexts as DocTree}  
\label{sec:DocTree}
Representing long contexts as a hierarchical DocTree forms the foundation of the ToM framework.
This transformation converts a flat, sequential document into a structured, tree-based representation, better reflecting the complexity of human cognition and enabling more organized, hierarchical reasoning.
The construction of the DocTree involves two key components: a \textit{Hierarchical Semantic Parser} that parses each chunk into an initial subtree, and a \textit{Bottom-up Tree Aggregation} process that merges these subtrees into a coherent hierarchy.

\noindent\textbf{Hierarchical Semantic Parser.}
Let \( D \) denote a long document, which is segmented into fixed-length token chunks \( \{c_1, c_2, c_3, \dots, c_n\} \), where \( n \) is the total number of chunks, and each \( c_i \) (\( 1 \leq i \leq n \)) contains a fixed number of tokens (e.g., 4k). Existing DCF approaches process \( \{c_i\} \) independently, treating each \( c_i \) as a discrete unit without considering its internal semantic structure.

In contrast, ToM introduces a semantic hierarchy within each chunk \( c_i \) by leveraging a Hierarchical Semantic Parser (HSP), a 3B-scale LLM distilled from GPT-4o. The HSP processes each chunk \( c_i \) and organizes it into a set of structured semantic subtrees \( \{t_{i,1}, t_{i,2}, \dots, t_{i,m}\} \), where \( m \) is the number of subtrees within \( c_i \). Each subtree \( t_{i,j} \) (\( 1 \leq j \leq m \)) encodes hierarchical relationships, such as parent-child links between main headings and subheadings. For simpler chunks, \( m = 1 \), resulting in a single hierarchical tree, while for more complex chunks, \( m > 1 \), forming a semantic forest.
As shown in Figure~\ref{fig:Tree Construct}, each chunk is processed by the HSP to extract its internal hierarchy and transform the hierarchical relationships into a subtree containing structured information such as Title, Keywords, Summary , and Context from the original document.
This hierarchical organization allows \( D \) to be represented as a collection of semantic subtrees \( \{t_{i,j}\} \), facilitating deeper structural understanding. 
Detailed implementation and quality assessment of HSP are discussed in Appendix~\ref{quality_of_HSP}, with prompts and illustrative cases provided in Appendix~\ref{Prompt_and_case}.

\noindent \textbf{Bottom-up Tree Aggregation.}  
After individual subtrees \( \{t_{i,j}\} \) are generated from each chunk \( c_i \), the next step is bottom-up aggregation to construct the hierarchical DocTree. This step involves embedding, clustering, and summarizing the root nodes of each subtree, recursively building higher-level summaries to form the complete DocTree.

The initial layer for DocTree, denoted as \( \mathcal{L}_0 \), is composed of the root nodes of all parsed subtrees:
\[
\mathcal{L}_0 = \bigcup_{i=1}^{n} \bigcup_{j=1}^{m_i} \{\text{SubtreeRoot}(t_{i,j})\},
\]
where \( n \) is the number of chunks, \( m_i \) is the number of subtrees within chunk \( c_i \), and \( \text{SubtreeRoot}(t_{i,j}) \) represents the root node of the \( j \)-th subtree \( t_{i,j} \). At this stage, \( \mathcal{L}_0 \) contains only the root nodes of all subtrees, while their child nodes remain encapsulated within their respective subtrees.

To construct the higher layers \( \mathcal{L}_k \), we perform the following steps iteratively:

\begin{enumerate}
    \item \textbf{Embedding}: Each root node \( r \in \mathcal{L}_{k-1} \) is embedded as a vector \( e_r \) using a pre-trained embedding model.
    \item \textbf{Clustering}: A clustering algorithm groups the embeddings into clusters \( \mathcal{I}_{k-1} = \{I_{k-1}^1, I_{k-1}^2, \dots, I_{k-1}^t\} \), where \( t \) is the number of clusters at layer \( k-1 \).
    \item \textbf{Summarization}: For each cluster \( I_{k-1}^j \in \mathcal{I}_{k-1} \), the evaluated LLM generates a summarized node \( s_k^j \), where:
    \[
    \mathcal{L}_k = \{s_k^1, s_k^2, \dots, s_k^t\}.
    \]
\end{enumerate}

Edges are created between the root nodes in \( I_{k-1}^j \) at layer \( \mathcal{L}_{k-1} \) and their corresponding summary node \( s_k^j \) at layer \( \mathcal{L}_k \).

This bottom-up aggregation is recursive and continues until the number of clusters stabilizes, resulting in the top layer \( \mathcal{L}_K \), which contains a single root node or a small number of high-level summary nodes representing the entire context. The final hierarchical DocTree can be represented as:
\[
\mathcal{T} = \{\mathcal{L}_0, \mathcal{L}_1, \dots, \mathcal{L}_K\}.
\]
The key steps for constructing DocTree are outlined in Algorithm 1.
The complete DocTree is constructed by organizing lower-level subtrees beneath higher-level summaries.
By hierarchically organizing information during aggregation, the lower levels of the DocTree retain detailed information, while the higher levels provide condensed summaries. The DocTree effectively captures both local and global relationships across subtrees, facilitating reasoning from local to global.

\begin{algorithm}
\hrule
\vspace{1.5mm}
\textbf{\small Algorithm 1: DocTree Construction Process}
% \label{Alg:construction}
\vspace{1.5mm}
\hrule
\small    
\fontsize{8}{10}\selectfont
\SetAlgoLined
\DontPrintSemicolon
\SetAlgoVlined
\SetAlgoSkip{1.5mm}  % 统一调整行距
\vspace{1.5mm}
\KwIn{Long context \( C \), Query \( q \)}
\KwOut{Constructed DocTree \( \mathcal{T} \)}
\BlankLine
\textbf{Step 1: Subtree Generation} \;
Split \( C \) into fixed-length token chunks \( \{c_1, c_2, \dots, c_n\} \)\ ;
\ForEach{chunk \( c_i \)}{
    Use the Hierarchical Semantic Parser to identify semantic structures\;
    Generate subtrees \( \{t_{i,1}, t_{i,2}, \dots, t_{i,m_i}\} \) for each chunk \( c_i \)\ ;
}
\BlankLine
\textbf{Step 2: Bottom-Up Tree Aggregation} \;
Initialize the bottom layer:
\[
\mathcal{L}_0 = \bigcup_{i=1}^n \bigcup_{j=1}^{m_i} \{\text{SubtreeRoot}(t_{i,j})\}
\]
\For{\( k = 1, 2, \dots \) until the number of clusters stabilizes} {
    \ForEach{node \( r \in \mathcal{L}_{k-1} \)}{
        Compute embeddings \( e_r \)\ ;
    }
    Cluster embeddings into groups \( \mathcal{I}_{k-1} = \{I_{k-1}^1, I_{k-1}^2, \dots, I_{k-1}^t\} \)\ ;
    Summarize each cluster \( I_{k-1}^j \) into a parent node \( s_k^j \)\ ;
    Update \( \mathcal{L}_k = \{s_k^1, s_k^2, \dots, s_k^t\} \)\ ;
}
\BlankLine
\textbf{Step 3: Final DocTree Construction} \;
Merge all aggregated nodes \( \mathcal{L}_K \) into the unified DocTree \( \mathcal{T} \), maintaining hierarchical relationships\ ;
\BlankLine
\Return{\textnormal{DocTree} \( \mathcal{T} \)}

\vspace{2mm}
\hrule
\end{algorithm}

\noindent \textbf{DocTree Compression.}
Short-cutting techniques reduce computational overhead in tree-based reasoning, particularly for large-scale DocTrees with millions of tokens.
Drawing inspiration from Retrieval-Augmented Generation methods, our approach integrates DocTree with retrieval techniques like BM25~\cite{robertson2009probabilistic} and embedding models, enabling the construction of a sparser, more concise DocTree. This process filters out irrelevant information, focusing on content closely aligned with the query, thereby improving both efficiency and relevance.
Algorithm 2 in Appendix~\ref{DocTree Construction Details} outlines the query-aware DocTree compression process, where the most relevant chunks are selected to build the DocTree. Constructing a DocTree from these aligned chunks incurs a small embedding cost while enhancing reasoning performance. Additionally, compressed DocTrees reduce token usage compared to those constructed from full documents, improving computational efficiency.

An example of DocTree construction is illustrated in Appendix~\ref{case_study}. A detailed analysis of the computational complexity involved in constructing the DocTree is also provided in Appendix~\ref{constructing_time}.

\subsection{Recursive MapReduce Reasoning}
\label{sec:Mapreduce}
ToM employs reasoning in a MapReduce style with two key steps: \textbf{Map} and \textbf{Reduce}.
These steps are applied recursively from the bottom up in the DocTree, propagating information and ensuring semantic coherence across the hierarchy.

In the Map step, rationales are generated in parallel from child nodes, while in the Reduce step, these rationales are aggregated across sibling nodes to resolve conflicts or reach consensus at the parent node. Each reasoning step produces a structured output, including key information, rationale, intermediate answer, and confidence estimates for conflict resolution, thereby enabling a more clear and interpretable reasoning process.
Map and Reduce are defined as follows:

\begin{enumerate}
    \item \textbf{Map:} \\ % 手动换行
    \textbf{Input:} For a leaf node, the input is \( \text{info}(\text{node}) \). For a non-leaf node, the input is \( \text{info}(\text{node}) \cup \text{Reduce}(\text{children}(\text{node})) \). \\
    \textbf{Operate:} Perform reasoning based on the current node's information. \\
    \textbf{Output:} \\\{key\_info, rationale, answer, confidence\}

    \item \textbf{Reduce:} \\ % 手动换行
    \textbf{Input:} Results of the Map step from sibling nodes at the same hierarchy level. \\
    \textbf{Operate:} Resolve conflicts and reach consensus among sibling nodes. \\
    \textbf{Output:} Aggregated \{key\_info, rationale, answer, confidence\} \\
    \textbf{Special Case:} Skip Reduce if \( |\text{siblings}| = 0 \).
\end{enumerate}

\begin{figure}[!t]
    \centering
    \includegraphics[width=1.\linewidth]{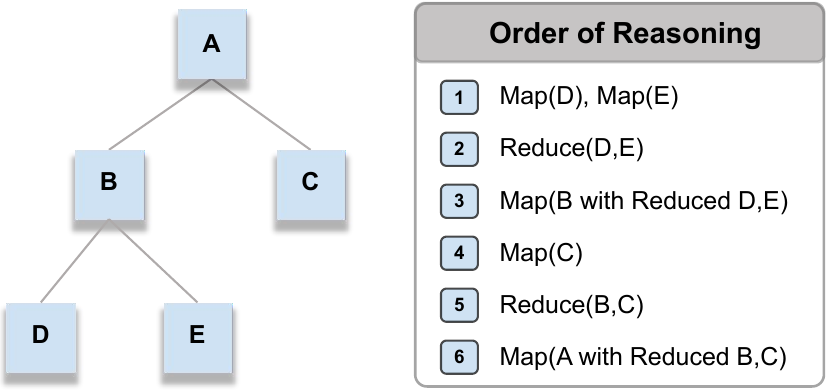}
    \caption{Illustration of the order of recursive reasoning. Nodes at the same hierarchy level, like D and E, can be processed in parallel for accelration.}
    \label{fig:reasoning_order}
\end{figure}
As illustrated in Figure~\ref{fig:reasoning_order}, recursive reasoning begins at the leaf nodes and propagates upward through Map and Reduce operations. Nodes at the same hierarchical level, such as D and E, are processed in parallel, enhancing computational efficiency. This combined application of Map and Reduce transforms the hierarchical DocTree into a structured reasoning framework, integrating local details into a cohesive global perspective.

For further details about Map and Reduce, see Appendix~\ref{Prompt_and_case} for the prompt, Appendix~\ref{case_study} for example cases, and Appendix~\ref{reasoning_time_test} for a comparison of ToM's average reasoning time.

\begin{table*}[!t]
\centering
\small
\resizebox{\textwidth}{!}{
\begin{tabular}{lcccccccccccccc}
\toprule
\multirow{2}{*}{\textbf{}} & & \multicolumn{2}{c}{\textbf{HQA (9.1k)}} & & \multicolumn{2}{c}{\textbf{2Wiki (4.9k)}} & & \multicolumn{2}{c}{\textbf{MuSi (11.2k)}} & & \multicolumn{2}{c}{\textbf{Inf.QA (192k)}} & & \textbf{Inf.MC (184k)} \\
\cmidrule(lr){3-4} \cmidrule(lr){6-7} \cmidrule(lr){9-10} \cmidrule(lr){12-13} \cmidrule(lr){15-15}
 & & \textbf{F1} & \textbf{RL} & & \textbf{F1} & \textbf{RL} & & \textbf{F1} & \textbf{RL} & & \textbf{F1} & \textbf{RL} & & \textbf{Acc} \\
\midrule
\multicolumn{15}{c}{\textit{\textbf{LLMs with Full Document}}} \\
\llamalogowithtext & & 21.66& 21.29 & & 21.48 & 21.41 & & 11.33 & 11.14 & & 8.32 & 8.09 & & 32.00 \\
\qwenlogowithtext & & 23.99 & 23.87 & & 21.75 & 21.67 & & 10.52 & 10.34 & & 12.47 & 12.12 & & 36.00 \\
\deepseeklogowithtext & & 41.41 & 41.02 & & 37.20 & 37.05 & & 32.88 & 32.77 & & 16.89 & 16.09 & & 47.00 \\
\gptlogowithtext & & 55.21 & \ 54.95 & & \ 44.28 & 44.29 & & \ 40.26 & \ 39.95 & & 13.45 & 12.99 & & 45.00 \\
\deepseekRlogowithtext & & 55.70 & 55.99 & & 59.59 & 59.24 & & 40.12 & 39.46 & & 12.58 & 13.09 & & 57.00 \\
\midrule
\multicolumn{15}{c}{\textit{\textbf{LLMs Equipped with RAG}}} \\
\llamalogowithtext & & 32.69 & 32.38 & & 25.39 & 25.47 & & 12.00 & 12.77 & & 13.32 & 14.99 & & 40.00 \\
\qwenlogowithtext & & 35.14 & 36.07 & & 22.58 & 23.78 & & 23.78 & 23.60 & & 19.65 & 19.94 & & 41.00 \\
\deepseeklogowithtext & & 42.33 & 41.17 & & 40.95 & 40.86 & & 28.26 & 29.18 & & 17.48 & 18.69 & & 53.00 \\
\gptlogowithtext & & 53.73 & 54.19 & & 54.99 & 53.79 & & 35.64 & 35.43 & & 26.03 & 25.47 & & 65.00 \\
\deepseekRlogowithtext & & 51.84 & 51.03 & & 56.12 & 56.97 & & 36.28 & 37.15 & & 23.14 & 24.49 & & 62.00 \\
\midrule
\multicolumn{15}{c}{\textit{\textbf{Divide-and-Conquer Frameworks}}} \\
\llamalogowithLA & & 45.31 & 45.68 & & 43.72  & 42.06 & & 26.81 & 25.60 & & 23.33 & 23.69 & & 62.00 \\
\qwenlogowithLA & & 47.19 & 48.38 & & 48.94  & 47.98 & & 28.16 & 27.60 & & 25.68 & 26.12 & & 65.00 \\
\deepseeklogowithLA & & 57.04 & 52.71 & & 56.87  & 55.23 & & 42.99 & 41.20 & & 29.63 & 27.12 & & 69.00 \\
\gptlogowithLA & & 55.25 & 52.30 & & 54.53 & 51.16 & & 45.44 & 44.03 & & 38.00 & 33.34 & & 72.00 \\
\deepseekRlogowithLA & & 59.72 & 58.66 & & \underline{61.90} & \underline{62.00} & & 45.41 & 44.86 & & 29.16 & 30.20 & & 76.00 \\
% \qwenqwqlogowithMoD & & 49.01 & 49.34 & & 48.42 & 48.42 & & 35.91 & 35.62 & & 31.37 & 31.68 & & 74.00  \\
\llamalogowithToM & & 50.40 & 52.71 & & 49.36  & 50.15 & & 30.89 & 31.20 & & 27.80 & 27.12 & & 68.00 \\
\qwenlogowithToM & & 52.19 & 52.08 & & 50.22  & 51.36 & & 28.29 & 29.50 & & 30.17 & 31.58 & & 71.00 \\
\deepseeklogowithMoD & & 60.87 & 60.43 & & 58.20 & 58.20 & & \textbf{50.15} & \textbf{50.04} & & \underline{38.60} & \underline{38.14} & & 77.00 \\
    \gptlogowithMoD & & \underline{61.07} & \underline{60.73} & & 59.31 & 59.55 & & \underline{47.27} & \underline{47.27} & & \textbf{41.17} & \textbf{46.04} & & \textbf{85.00} \\
\deepseekRlogowithMoD & & \textbf{61.91} & \textbf{61.55} & & \textbf{63.33} & \textbf{63.25} & & 44.74 & 44.86 & & 34.54 & 34.61 & & \underline{80.00} \\
\bottomrule
\end{tabular}
}
\caption{Long document reasoning performance (\%) on Longbench and InfiniteBench, with average token lengths from 4.9k to 192k. 100 samples are randomly selected for Inf.QA and Inf.MC, while the full sets are used for the other tasks. Bold denotes the best performance, and underlined results indicate runner-ups.}
\label{tab:main_table}
\end{table*}

\section{Experiment}
\subsection{Experiment setup}

\noindent  \textbf{Models.} 
Open-source LLMs with 70B+ parameters, including Qwen2.5-72B-Instruct~\cite{qwen2025qwen25technicalreport}, Llama3-70B-Instruct, DeepSeek-V3~\cite{deepseekai2024deepseekv3technicalreport},  and GPT-4o-2024-05-01. Additionally, reasoning model Deepseek-R1 is also evaluated for comparison.

\noindent  \textbf{Benchmarks.}
We evaluate the performance on \textsc{HotpotQA}, \textsc{2WikiMQA}, and \textsc{Musique} from LongBench~\cite{bai2024longbench}.
For more challenging tasks, we test on the Question-Answering (Inf.QA) and Multi-Choice (Inf.MC) tasks from InfiniteBench~\cite{zhang-etal-2024-bench}, where the average input length is 190k tokens. This poses a significant challenge for LLMs to process, understand, and reason over ultra-long contexts. Additional comparisons on Long In-Context Learning and Long Dialogue History Understanding tasks are provided in Appendix~\ref{additional_exp}.

\noindent \textbf{Baselines.}
We compare ToM with the following baselines: \textbf{(i) Reasoning on Full Document}: Query and full documents are concatenated for reasoning. Contexts exceeding the window size are truncated. \textbf{(ii) LLM enhanced with RAG}: For the RAG baseline, we perform chunking using a fixed number of tokens: 1k for Longbench and 4k for InfiniteBench. We adopt BGE-M3 with an 8k token window as the retriever and select the Top 5 chunks for augmented generation. We report the results of 70B+ LLMs, including Qwen2.5-72B-Instruct, Llama3-70B-Instruct, Deepseek-V3, GPT-4o, and Deepseek-R1. Additionally, the Appendix~\ref{compare with RAGs} provides an in-depth comparative analysis between RAG and ToM. \textbf{(iii) Divide-and-Conquer Frameworks}: Recent divide-and-conquer framework LongAgent~\cite{zhao-etal-2024-longagent} is evaluated. LongAgent utilizes multiple member agents, each responsible for processing assigned context chunks. Following the chunking process, a leader judge agent synthesizes the final answer through a multi-round discussion. For \textsc{\textsc{Inf.QA}} and \textsc{Inf.MC}, the chunk length is set to 4k tokens, while for other datasets in Longbench, it is configured to 1k tokens. Details about ToM, including the embedding technique, chunking size, and clustering algorithm, are provided in Appendix~\ref{sec:appendix_tom_components_and_implementation}.

\noindent \textbf{Metrics.} We use F1-score and RougeL score to evaluate question-answering performance and Accuracy for multiple-choice tasks.

\subsection{Main Results}
The overall results in Table~\ref{tab:main_table} highlight ToM's performance compared to existing methods. We analyze the results and present the following findings:

\textbf{Challenge Exists in Long Document Reasoning.} LLMs still face significant challenges in long-context reasoning, ranging from 10k tokens to ultra-long scenarios. Qwen2.5-72B-Instruct performs poorly on 10k-token documents like \textsc{HotpotQA} and \textsc{MuSique}, with F1-scores between 10.52\% and 23.99\%. DeepSeek-V3 outperforms Qwen by 17.4\% on \textsc{HotpotQA} and 15.5\% on \textsc{2WikiMQA} with full-document context.

This challenge becomes more apparent when handling ultra-long contexts exceeding 100k tokens. Tasks like \textsc{Inf.QA} and \textsc{Inf.MC}, averaging 190k tokens, present significant challenges for all LLMs.
Llama3-70B-Instruct and Qwen2.5-72B-Instruct score 8.32\% and 12.47\%, respectively, while GPT-4o achieves 13.45\%, and DeepSeek-V3 leads with 16.89\% on \textsc{Inf.QA}.
Even DeepSeek-R1, with state-of-the-art performance still struggles with ultra-long contexts, achieving only 12.58\% on \textsc{Inf.QA}. These results highlight the limitations of current LLMs and the need for improved methods to handle such extensive contexts.

\textbf{RAG Benefits in Long Document Reasoning.}
On short document reasoning tasks, with the exception of GPT-4o, most LLMs show significant gains from RAG techniques. For example, on \textsc{HotpotQA}, Llama3-70B-Instruct's F1 score increases from 21.66\% to 32.69\%, and Qwen2.5-72B-Instruct's from 23.99\% to 35.14\%. Interestingly, DeepSeek-R1 sees a slight drop, from 55.70\% to 51.84\%, suggesting that its strong in-context performance may be disrupted by the fragmentation introduced by chunking.

For ultra-long documents, RAG provides a straightforward solution by helping LLMs focus on the most relevant chunks. With RAG, Llama3-70B-Instruct and Qwen2.5-72B-Instruct achieve notable improvements on \textsc{Inf.QA}, reaching F1 scores of 13.32\% and 19.65\%, respectively, while GPT-4o sees a boost to 26.03\%.
Similarly, on \textsc{Inf.MC}, all LLMs benefit from RAG, with performance gains ranging from 5.0\% to 20.0\%.

However, while RAG improves performance in long-context reasoning, it still faces two key limitations: 1) Although RAG improves the recall of relevant chunks, it does not ensure that the retrieved chunks are truly useful for reasoning, often introducing irrelevant information that adds noise to the process. 2) RAG-enhanced LLMs reason over flat-organized chunks but often sacrifice logical coherence due to their reliance on similarity-based rankings, leaving interconnections between chunks underutilized.
These limitations constrain the upper bound of RAG-based methods, highlighting the need for a more fine-grained reasoning approach that fully leverages the chunks and improves information aggregation.

% However, while RAG improves chunk recall, it does not guarantee high precision.
% QA tasks often require accurate, fine-grained reasoning that can be disrupted by irrelevant or fragmented retrievals.
% This mismatch limits RAG’s effectiveness and highlights the need for better alignment between retrieval and generation.

\textbf{DCF Outperform One-Step Reasoning in Long Contexts Scenarios.}
Divide-and-conquer methods consistently outperform one-step reasoning approaches.
LongAgent significantly improves performance across all evaluated LLMs. On \textsc{HotpotQA}, LongAgent boosts DeepSeek-R1 to 59.72\% F1, surpassing its RAG baseline by 7.88\%. Llama3-70B-Instruct and Qwen2.5-72B-Instruct see even larger gains of 12.62\% and 12.05\%, respectively.
In ultra-long document reasoning, LongAgent also proves effective. With DeepSeek-R1, it achieves 29.16\% F1 on \textsc{Inf.QA} and 76.00\% accuracy on \textsc{Inf.MC}. GPT-4o performs best on \textsc{Inf.QA} at 38.00\%, while maintaining 72.00\% on \textsc{Inf.MC}. 
% LA also substantially improves Llama3-70B-Instruct and Qwen2.5-72B-Instruct, reaching 23.33\% and 25.68\% on \textsc{Inf.QA}, respectively. 
These results underscore the advantage of divide-and-conquer strategies in complex, long-context reasoning.

LongAgent enhances long-context reasoning performance by breaking long documents into smaller components for parallel reasoning and employing a Leader LLM to resolve conflicts and refine results across multiple turns. Though effective for local reasoning, LongAgent processes chunks in isolation, limiting its ability to capture long-range dependencies and increasing the risk of conflicts, thus capping its potential.

% \subsection*{ToM Achieves Significant Performance Gains Compared to Other Divide-and-Conquer Schema}

\textbf{ToM Achieves Significant Performance Gains.} With GPT-4o, ToM achieves state-of-the-art performance, reaching 41.17\% F1 on \textsc{Inf.QA} and 85.0\% accuracy on \textsc{Inf.MC}.
When paired with Qwen2.5-72B-Instruct and DeepSeek-V3, ToM also demonstrates strong performance, achieving 30.17\% and 38.60\% F1 on \textsc{Inf.QA}, and 71.00\% and 77.00\% accuracy on \textsc{Inf.MC}, respectively, consistently outperforming their LongAgent counterparts.
Compared to RAG-based methods, ToM with GPT-4o improves performance by 15.14\% on \textsc{Inf.QA} and 20.0\% on \textsc{Inf.MC}. Relative to LongAgent with GPT-4o, it achieves gains of 11.97\% in F1 and 13.0\% in accuracy, highlighting ToM's advantage in ultra-long document reasoning.

ToM’s effectiveness stems from its recursive tree-based MapReduce framework. In the Map phase, relevant information is extracted from chunks; in the Reduce phase, results are merged, conflicts resolved, and refined using confidence scores. This structured approach reduces noise and enhances integration, making it well-suited for long context reasoning tasks.

Compared to RAG, ToM performs fine-grained reasoning on informative content identified by HSP within each chunk, fully leveraging relevant content while filtering out noise based on confidence estimates. 
Compared to LongAgent, its tree-based MapReduce framework enables more effective conflict resolution and enhances fact aggregation from local to global through the DocTree structure, thereby achieving promising performance gains.

\textbf{Equip ToM with Reasoning Models.} We evaluate both LongAgent and ToM using DeepSeek-R1. On shorter tasks like \textsc{HotpotQA} and \textsc{2WikiMQA}, DeepSeek-R1 combined with either framework outperforms other models, with ToM achieving 61.91\% and 63.33\% F1, respectively. As document length increases, ToM with GPT-4o surpasses DeepSeek-R1 on ultra-long tasks such as \textsc{Inf.QA} and \textsc{Inf.MC}.
While the long chains of thought produced by R1 benefit understanding in shorter contexts, they also introduce overthinking and hallucinations during reasoning, allowing incorrect ideas to propagate upward and ultimately impairing overall reasoning.

 % This suggests that different models may be optimal for different document lengths, with reasoning-focused models excelling in moderate-length scenarios while more balanced models like GPT-4o may be preferable for ultra-long contexts.

\section{Ablation Study}
In this section, we discuss the contributions of key components, analyze efficiency, and examine the impact of compression, with additional ablations on chunk size provided in Appendix~\ref{impact_of_chunk_size}.

\begin{table}[!b]
\centering
\small
\resizebox{\columnwidth}{!}{
\begin{tabular}{ccccccc}
\toprule
\textbf{Aggre.} & \textbf{Conf.} & \textbf{HQA} & \textbf{2Wiki} & \textbf{MuS} & \textbf{Inf.QA} & \textbf{Inf.MC} \\
\midrule
\ding{52} & \ding{52} & \textbf{61.1} & \textbf{59.3} & \textbf{47.3} & \textbf{38.6} & \textbf{85.0}  \\
\ding{55} & \ding{52} & 55.8 & 53.0 & 42.5 & 36.6 & 79.0  \\
\ding{52} & \ding{55}  & 56.5 & 51.4 & 39.3 & 31.7 & 78.0  \\
\bottomrule
\end{tabular}
}
\caption{Effect of the in-context confidence measure and bottom-up aggregation with DeepSeek-V3.}
\label{tab:component}
\end{table}

\noindent \textbf{Contribution of Key Components.} 

We conduct ablations on bottom-up aggregation and the confidence measure, with results shown in Table~\ref{tab:component}.
Removing the in-context confidence measure leads to substantial performance degradation across all datasets (–6.9\% on \textsc{Inf.QA}, –7.0\% on \textsc{Inf.MC}). In the tree-based MapReduce reasoning pattern, the reduce phase is essential for merging viewpoints and resolving conflicts, which is key to strong performance. Divide-and-conquer reasoning often encounters local conflicts, and confidence scores help guide resolution for more consistent results.

Removing bottom-up aggregation results in consistent performance drops (–2.0\% on \textsc{Inf.QA}, –6.0\% on \textsc{Inf.MC}). It contributes by providing global context through hierarchical summaries, which are difficult to infer solely from detailed subtree information. By integrating low-level details with high-level summaries, ToM enables more effective reasoning over the DocTree.

\noindent \textbf{Efficiency Analysis.}
ToM introduces the Hierarchical Semantic Parser (HSP) to capture the internal hierarchy within each chunk. While adding some computational cost, the impact is minimal. 
First, we use a lightweight 3B-scale LLM for parsing, which is more affordable than relying heavily on expensive GPT API calls. Second, we leverage technologies like vLLM to enable parallel processing, further optimizing efficiency.
% HSP processes an average of 1.9 chunks of 4k tokens per second on a single A100 GPU, demonstrating its efficiency.
The average DocTree construction time across different stages and input lengths is provided in Appendix~\ref{constructing_time}.
ToM requires fewer LLM calls than LongAgent, making 4.2k calls on 100 \textsc{Inf.QA} samples compared to LongAgent’s 6.3k.

\begin{figure}[!ht]
    \centering
    \includegraphics[width=1\linewidth]{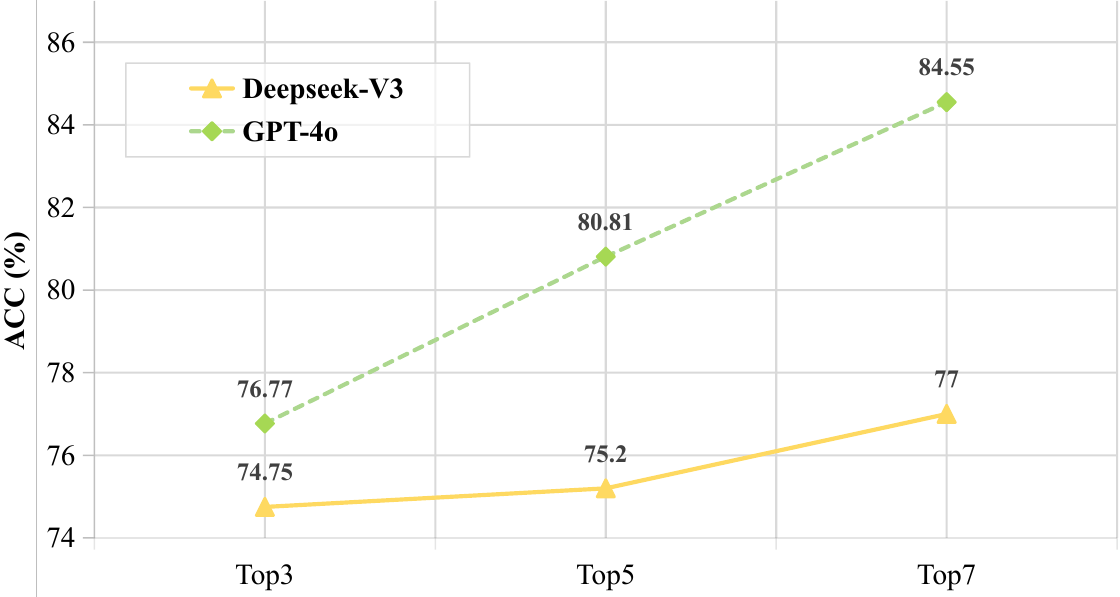}
    \caption{Effect of the number of selected chunks.}
    \label{fig:Compression_Ratio}
\end{figure}

\noindent \textbf{Compression Impact.}
We compress Doctree by selecting relevant chunks for construction, with the reasoning performance versus the number of selected chunks on \textsc{Inf.MC} shown in Figure~\ref{fig:Compression_Ratio}. There is a trade-off between reasoning performance and computational costs: while increasing the number of chunks raises overhead, it also improves performance. Notably, as the number of chunks increases from Top-3 to Top-7, reasoning performance grows steadily for GPT-4o, highlighting the benefits of adding more context despite the additional costs.

\section{Conclusion}
In this paper, we introduced ToM, a tree-oriented MapReduce framework for effective long document reasoning with large language models. By leveraging tree-based representations and recursive MapReduce reasoning, ToM enhances fact utilization and conflict resolution across extended documents.
Extensive experiments show that ToM achieves significant performance gains over existing divide-and-conquer and RAG-enhanced frameworks, highlighting the potential of tree-based frameworks to overcome the limitations of current LLMs in long-context reasoning and paving the way for more effective, scalable solutions.
% \clearpage
\section*{Limitations}
ToM significantly enhances long-context reasoning through its novel tree-based MapReduce framework. However, the construction of the DocTree, which involves hierarchical semantic parsing and bottom-up aggregation, introduces computational overhead, particularly when processing ultra-long documents. Although the query-aware DocTree compression improves efficiency, it may inadvertently omit intermediate information that is essential for complex multi-hop reasoning. Additionally, the granularity of text chunking and the potential for error propagation from earlier analytical stages can affect the overall performance of the framework.
Future research includes extending ToM’s capabilities to handle multimodal documents for more comprehensive long-context understanding.

\section*{Acknowledgments}
This work was supported by the National Natural Science Foundation of China (No. 62306216), and the Technology Innovation Program of Hubei Province (No. 2024BAB043). Hai Zhao was funded by The Major Program of Chinese National Foundation of Social Sciences under Grant ‘The Challenge and Governance of Smart Media on News Authenticity’ [No. 23\&ZD213]. Yujiu Yang was supported by the National Key Research and Development Program of China (No. 2024YFB2808903) and the  research grant No. CT20240905126002 of the Doubao Large Model Fund.
\bibliography{main}
\clearpage
\newpage

\appendix
\section*{Appendix}
This appendix provides supplementary material to further elaborate on the \method~framework. Section~\ref{sec:appendix_additional_eval} presents additional performance evaluations of \method~on long-context QA tasks. Section~\ref{sec:appendix_tom_components_and_implementation} offers implementation details and an in-depth analysis of \method's key components, including its computational complexity and chunking strategy, the Hierarchical Semantic Parser's implementation and quality, and the DocTree construction process covering aggregation and compression techniques. Finally, Section~\ref{sec:appendix_tom_methodology_deep_dive} provides a deeper dive into \method's methodology through a detailed comparison with RAG, an exposition of its core operational prompts, and illustrative case studies.

% --- SECTION A: Additional Performance Evaluations ---
\section{Additional Performance Evaluations}
\label{sec:appendix_additional_eval}

This section extends the performance evaluations presented in the main paper, showcasing \method's effectiveness on further challenging long-context question answering tasks.

\subsection{Performance on User Guide QA and Dialogue History QA}
\label{additional_exp}
In addition to HQA, 2Wiki, MuSi, Inf.QA, and Inf.MC, we also evaluated \method~on additional QA tasks: the Long In-context Learning task—User Guide QA and the Long-dialogue History Understanding task—Dialogue History QA. The experimental results in Table~\ref{additional_tasks_appendix} demonstrate that on User Guide QA, our \method~approach achieved an accuracy of 65.0\%, outperforming the baseline GPT-4o model by 12.5 percentage points (from 52.5\%) and GPT-4o+LongAgent by 5.0 percentage points (from 60.0\%). Similarly, on Dialogue History QA, \method~reached 68.4\% accuracy, surpassing the baseline GPT-4o by 15.8 percentage points (from 52.6\%) and GPT-4o+LongAgent by 5.2 percentage points (from 63.2\%).

\begin{table}[!b]
\centering
\small
\resizebox{\columnwidth}{!}{%
\begin{tabular}{lcc}
\toprule
\textbf{} & \textbf{User Guide QA}  & \textbf{Dialogue History QA} \\
\midrule
GPT-4o & 52.5 & 52.6 \\
+ LongAgent & 60.0 & 63.2 \\
+ \method & \textbf{65.0} & \textbf{68.4} \\
\bottomrule
\end{tabular}
}
\caption{Accuracy on User Guide and Dialogue History.}
\label{additional_tasks_appendix} % Renamed label for uniqueness
\end{table}

These statistically significant and consistent performance improvements across diverse long-context tasks provide empirical evidence for \method's robust generalizability. The observed enhancements can be attributed to the fundamental architectural advantages of our framework: specifically, the tree-oriented MapReduce methodology enables hierarchical document structuring through semantic parsing and bottom-up aggregation, preserving critical semantic relationships that would otherwise be lost in traditional approaches. Unlike conventional RAG methods that rely on sequential reasoning or divide-and-conquer frameworks that process chunks in isolation, \method's recursive reasoning mechanism explicitly maintains parent-child relationships between document segments and facilitates long-range information flow across the entire DocTree. This structural coherence is particularly beneficial for dialogue-based tasks, as evidenced by the substantial 15.8 percentage point improvement on Dialogue History QA, where maintaining contextual continuity across multiple conversational turns is essential for accurate comprehension and reasoning. The experimental findings thus substantiate our hypothesis that tree-based reasoning offers a more effective paradigm for complex long-context understanding tasks that require sophisticated multi-hop inference capabilities.

\FloatBarrier
% --- SECTION B: Implementation Details and Component Analysis of ToM ---
\section{Implementation Details and Component Analysis of \method}
\label{sec:appendix_tom_components_and_implementation}

This section delves into specific implementation details and characteristics of the \method~framework's key components. We analyze its computational complexity alongside the chunking strategy, detail the training, implementation, and quality of the Hierarchical Semantic Parser, and discuss the DocTree construction process, focusing on aggregation and compression techniques along with their implications.

\subsection{Computational Complexity}
\label{constructing_time}
When comparing our approach, \method, to traditional methods like RAG or divide-and-conquer frameworks, one noticeable difference is the additional tree construction step that \method~incorporates. The construction of the DocTree in \method~consists of three steps: 1) \textbf{embedding chunks before constructing}; 2) \textbf{Inferencing during hierarchical semantic parsing}; and 3) \textbf{bottom-up summarizing}.

Table~\ref{tab:constructing_time_appendix} details the average time for DocTree construction over varying document lengths.
\begin{table}[!b]
\centering
\small
\begin{tabular}{lcccc}
\toprule
\textbf{Time (Component)} & \textbf{10k} & \textbf{80k} & \textbf{120k} & \textbf{250k} \\
\midrule
Embedding & 0.6 & 1.9 & 2.7 & 5.3  \\
Inference (HSP) & 9.9 & 29.4 & 34.3 & 37.2  \\
Summarization & 8.0 & 27.0 & 28.1  & 32.9  \\
\midrule
Total Construction & 18.5 & 58.3 & 65.1 & 75.4  \\
\bottomrule
\end{tabular}
\caption{Average DocTree constructing time (s) within different stages on varying lengths.}
\label{tab:constructing_time_appendix}
\end{table}

From the experimental results shown in Table~\ref{tab:constructing_time_appendix}, it is clear that the construction time increases with document length. The total time required for constructing the DocTree increases steadily, with processing times of 18.5 seconds for 10k tokens and 75.4 seconds for 250k tokens.

\paragraph{Chunking Size} The chunk size for input documents was adapted based on the benchmark: input documents were segmented into chunks of 1,000 tokens for Longbench tasks and 8,000 tokens for InfiniteBench tasks.

\subsection{Reasoning Time on Ultra Long Document}
\label{reasoning_time_test}
Table~\ref{tab:reasoning_time_appendix} presents the average reasoning time compared to other methods.
\begin{table}[!htbp]
\centering
\small
\begin{tabular}{lccc}
\toprule
 & \textbf{RAG} & \textbf{LongAgent} & \textbf{Our \method} \\
\midrule
Time (s) & 5.4 & 236.8 & 145.0 \\
\bottomrule
\end{tabular}
\caption{Time costs on Inf.QA over 100k tokens.}
\label{tab:reasoning_time_appendix}
\end{table}

Turning to the reasoning performance presented in Table~\ref{tab:reasoning_time_appendix}, \method~achieves reasoning times of 145.0 seconds for 100k tokens. While RAG is faster (5.4s), LongAgent is slower (236.8s). Although the construction and reasoning time for \method~can be higher, our experimental evaluations (presented in the main paper and Section~\ref{sec:appendix_additional_eval}) show that the added time is justified by the significant improvements in performance, offering a better balance between efficiency and effectiveness.

\subsection{Impact of Chunk Size.}
\label{impact_of_chunk_size}
\begin{figure}[!t]
    \centering
    \includegraphics[width=0.95\linewidth]{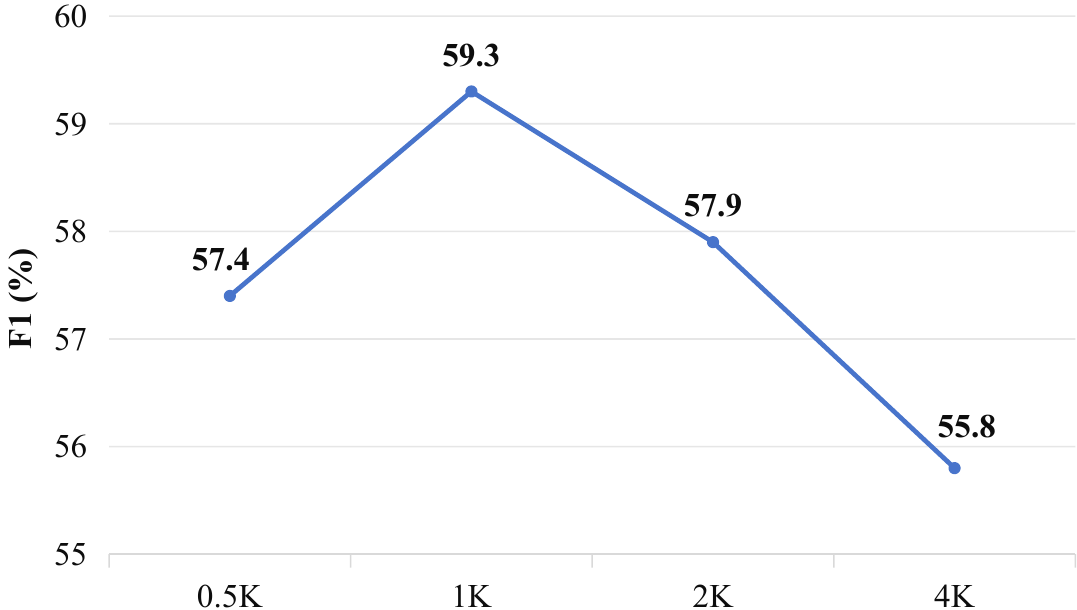}
    \caption{Effect of the chunk size.}
    \label{fig:chunk_size}
\end{figure}
We evaluate different chunk sizes (0.5k-4k tokens) on the 2WikiMQA dataset (average length 4.9k tokens) using ToM with GPT-4o. Results in Figure~\ref{fig:chunk_size} show that 1k tokens achieves optimal performance (59.3\% F1), while both smaller (0.5k, 57.4\% F1) and larger chunks (4k, 55.8\% F1) lead to degraded performance. Smaller chunks likely fragment semantic coherence, making it difficult to capture complete logical units. Conversely, larger chunks may reduce the effectiveness of the hierarchical structure by creating fewer but more complex nodes in the DocTree, potentially obscuring the parent-child relationships that facilitate effective reasoning. The optimal 1k chunk size creates a more balanced hierarchical structure with clearer semantic boundaries between nodes, allowing for more effective information organization for this particular dataset.

\subsection{Hierarchical Semantic Parser (HSP): Implementation and Quality Assessment}
\label{quality_of_HSP}
\paragraph{HSP Training and Implementation} The Hierarchical Semantic Parser (HSP) was trained using the Wiki727 dataset~\cite{koshorek-etal-2018-text} for initial data generation, with semantic chunking performed by GPT-4o~\cite{openai2024gpt4technicalreport}. To enhance operational efficiency, we distilled a 3B-scale model, Qwen2.5-3B-Instruct. This distillation involved fine-tuning the model on 18,000 query-response pairs generated by GPT-4o, employing a full-parameter supervised approach with an 8k context window. For parallel processing tasks related to the HSP and other components, vLLM~\cite{kwon2023efficient} was incorporated. Following the HSP parsing, subtrees were constructed from the parsed chunks using regular expressions to identify parent-child relationships; this method proved robust, yielding no matching failures across thousands of parsing instances.

\begin{figure}[!b]
    \centering
    \includegraphics[width=1.\linewidth]{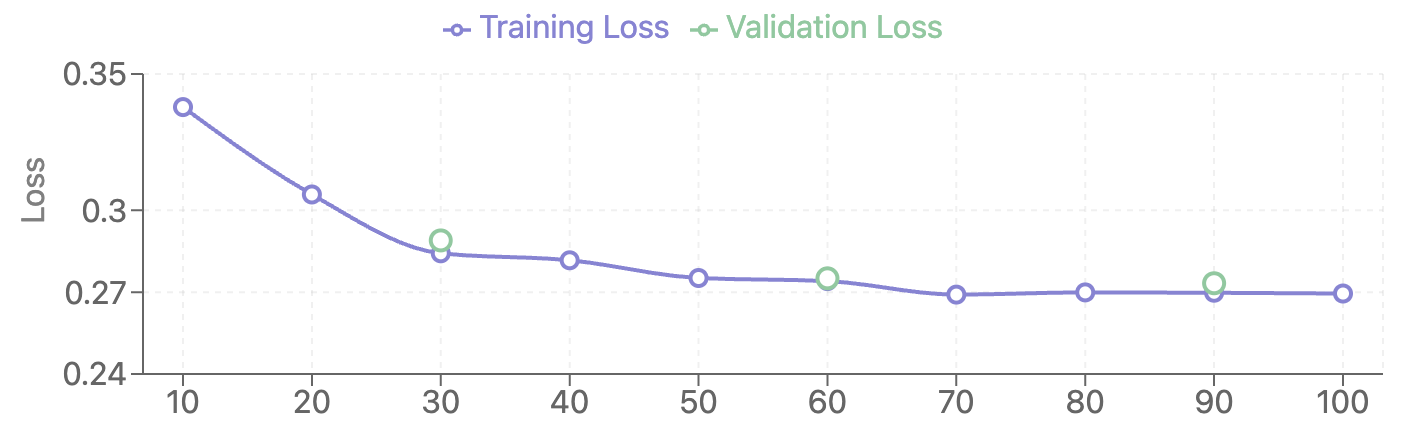}
    \caption{Loss curves of the HSP using the Qwen2.5-3B-Instruct.}
    \label{fig:loss_appendix}
\end{figure}

\paragraph{HSP Quality} Current LLMs can perform stable parsing without tuning and can achieve better performance after distillation. Figure~\ref{fig:loss_appendix} shows the loss curves for our HSP, indicating effective learning and stabilization of parsing quality.

\begin{figure}[!t]
    \centering
    \includegraphics[width=0.9\linewidth]{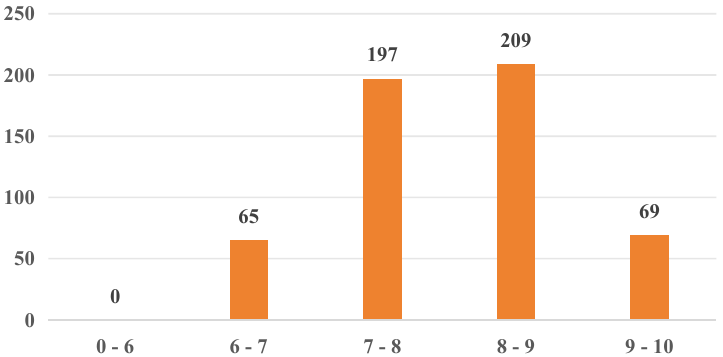}
    \caption{ Score distribution of 500 HSP samples evaluated by Claude-Sonnet-3.7, showing a mean score of 8.02.}
    \label{parse_quality}
\end{figure}

To further evaluate parsing reliability, we conducted a quality assessment adopting the LLM-as-judge approach using Claude-Sonnet-3.7 as an independent evaluator which is shown in Figure~\ref{parse_quality}. Our analysis of 500 parsed samples, scored on a scale of 0-10, reveals strong performance with no samples below 6.0, and the majority (406 samples, 81.2\%) scoring between 7.0-9.0. The highest concentration appears in the 8.0-9.0 range (209 samples), with an overall mean score of 8.02. This distribution confirms that our HSP consistently produces high-quality hierarchical structures, effectively capturing semantic relationships essential for the subsequent MapReduce reasoning process.

\subsection{DocTree Construction Details: Aggregation and Compression}
\label{DocTree Construction Details}
\paragraph{Recursive Summarization for Aggregation} In the recursive summarization process, which forms part of the bottom-up tree aggregation in DocTree construction, node clustering was performed using the Leiden~\cite{DBLP:journals/corr/abs-1810-08473} algorithm. The main evaluated LLM (as specified in the main paper's experimental setup) then performed the subsequent summarization steps for the clustered nodes.
\begin{algorithm}[!b]
\hrule
\vspace{1.5mm}\textbf{\small Algorithm 2: Query-Aware DocTree Compression}
% \caption{}
\label{Alg:compression_appendix}
\vspace{1.5mm}
\hrule
\small
\fontsize{8}{10}\selectfont
\SetAlgoLined
\DontPrintSemicolon
\SetAlgoVlined
\SetAlgoSkip{1.5mm}
\vspace{3mm}
\KwIn{Long context \( C \), Query \( q \), Retriever \( \mathcal{R} \), Selection scale \( k \)}
\KwOut{Compressed set of chunks \( \{c'_1, c'_2, \dots, c'_k\} \) to form DocTree \( \mathcal{T} \)}

\BlankLine
\textbf{Step 1: Embedding Generation} \;
Generate query embedding \( e_q \) for \( q \) using \( \mathcal{R} \)\;
Split \( C \) into fixed-length token chunks \( \{c_1, c_2, \dots, c_n\} \)\;
\ForEach{chunk \( c_i \)}{
    Generate embedding \( e_{c_i} \) for \( c_i \) using \( \mathcal{R} \)\;
    Compute cosine similarity \( \text{Sim}(e_{c_i}, e_q) \) for \( e_{c_i} \) and \( e_q \)\;
}
\BlankLine
\textbf{Step 2: Chunk Selection} \;
Select the top-\( k \) chunks based on similarity scores:
\[
\{c'_1, c'_2, \dots, c'_k\} = \text{TopKSim}(q, \{c_1, c_2, \dots, c_n\})
\]
\BlankLine
\textbf{Step 3: Build Compressed DocTree \( \mathcal{T} \) using \(\{c'_1, c'_2, \dots, c'_k\}\)}
\BlankLine
\textbf{Return:} Compressed DocTree \( \mathcal{T} \) (built from selected chunks)\;
\vspace{2mm}
\hrule
\end{algorithm}

\paragraph{DocTree Compression Algorithm} The DocTree compression strategy is crucial for managing computational resources. Algorithm 2 outlines our query-aware compression technique. For this mechanism, the parameter $k$ (number of selected chunks) was set to 7. Similarity scores, crucial for query-aware chunk selection during compression, were computed between query and node embeddings using the BGE-M3 model~\cite{DBLP:journals/corr/abs-2402-03216}.

\paragraph{Analysis of Information Loss during Compression}
Our DocTree compression strategy, which employs query-aware top-k selection using embeddings, does introduce a deliberate tradeoff between computational efficiency and information completeness. By selectively focusing on the most relevant chunks for DocTree construction, we reduce processing time while maintaining high performance on most reasoning tasks. However, this process may potentially exclude intermediate information that could be valuable for capturing certain long-range dependencies, particularly in cases requiring complex multi-hop reasoning across seemingly unrelated document sections.

To address this concern, we emphasize the flexible nature of our framework, which allows practitioners to adjust the compression parameters based on their specific requirements. For applications prioritizing maximum reasoning accuracy where computational resources are less constrained, compression can be made optional, permitting the construction of complete DocTrees that preserve all document information. 

Our experiments with varying compression levels indicate that increasing the number of selected chunks generally improves performance, with diminishing returns beyond a certain threshold. This suggests that while some information loss occurs during compression, our embedding-based selection effectively identifies most critical chunks, and the hierarchical structure of DocTree helps mitigate the impact of missing intermediate information by establishing connections between distant but semantically related chunks during the bottom-up aggregation process.

\FloatBarrier

% --- SECTION C: ToM Methodology: In-depth and Comparative View ---
\section{\method~Methodology: In-depth and Comparative View}
\label{sec:appendix_tom_methodology_deep_dive}

This section offers a more detailed exposition of \method's methodology. We provide an in-depth comparison with traditional Retrieval-Augmented Generation (RAG) to highlight \method's unique advantages, present the core prompts that guide its reasoning steps, and illustrate its operation through case studies.

\subsection{Comparison with Retrieval-Augmented Generation (RAG)}
\label{compare with RAGs}
\method~demonstrates superior performance across all benchmarks as evidenced by our experimental results. An intriguing aspect of this comparison is that \method~initially employs retrieval mechanisms similar to those in RAG for chunk selection, yet achieves significantly better outcomes in downstream reasoning tasks.

Retrieval-Augmented Generation (RAG) approaches have made significant contributions to question answering tasks by effectively enhancing recall—successfully retrieving chunks containing potentially relevant information. However, QA tasks inherently demand high precision in utilizing retrieved information to generate accurate answers. While RAG excels at gathering diverse relevant content, its flat concatenation of retrieved chunks and single-pass reasoning process presents challenges for complex queries requiring nuanced information integration. It lacks the ability to capture the logical structure between different information segments, struggles to establish connections between distantly related facts, and cannot effectively resolve contradictions that may appear across different chunks. Consequently, RAG's performance degrades significantly when handling queries that require integrating information from multiple document segments or performing multi-hop reasoning.

\begin{figure*}[!t]
    \centering
    \includegraphics[width=0.95\textwidth, height=0.95\textheight, keepaspectratio]{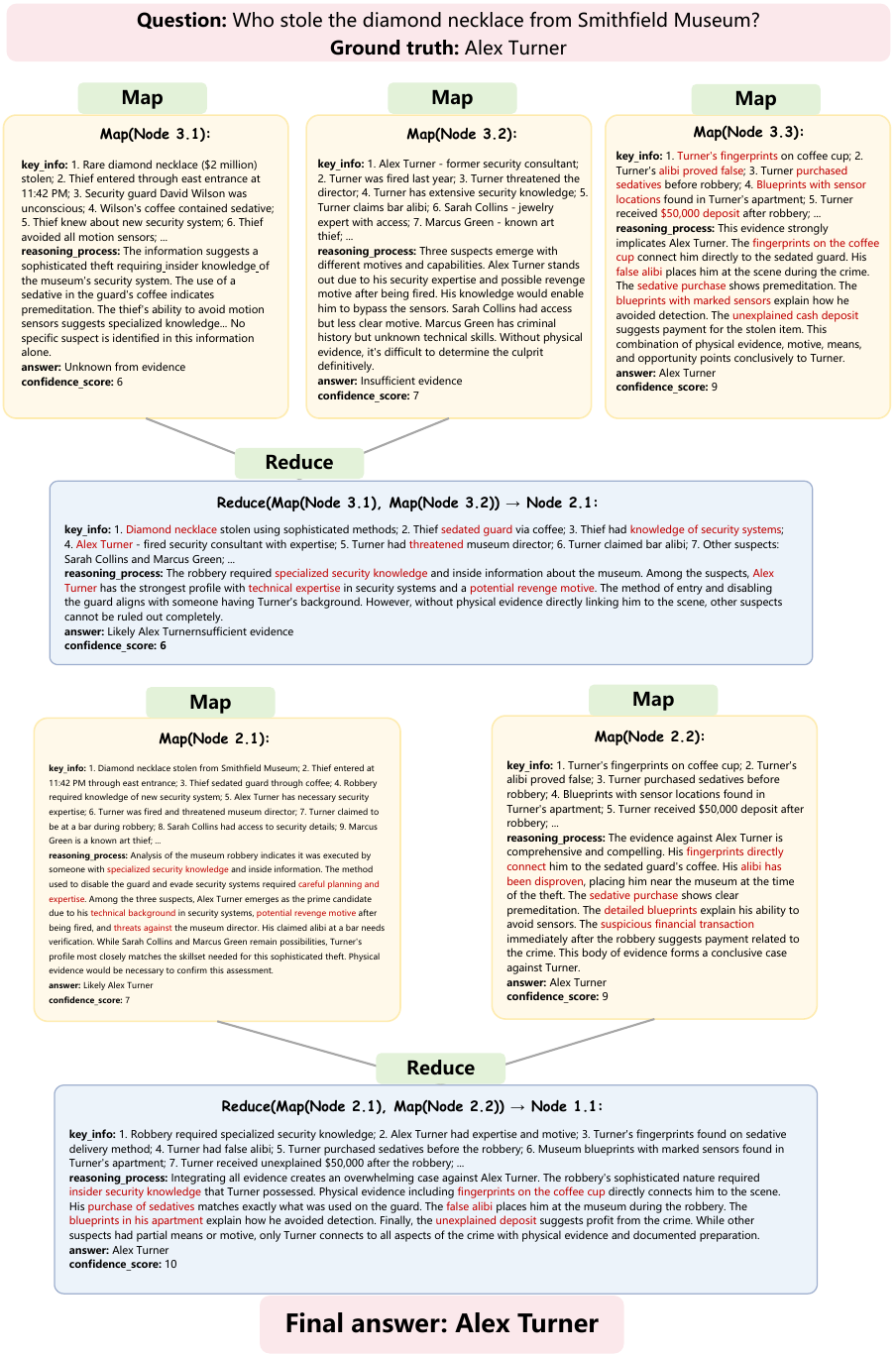}
    \caption{Case Study.}
    \label{fig:case_reasoning_appendix}
\end{figure*}

Our Tree-oriented MapReduce (\method) framework builds upon RAG's strong retrieval foundation while addressing this precision challenge through hierarchical processing. By organizing the same retrieved chunks into a structured DocTree, \method~enables a two-phase reasoning process that systematically improves precision. In the Map phase, \method~conducts fine-grained reasoning on individual sibling nodes in parallel, generating detailed rationales that capture essential supporting facts from each document segment. The subsequent Reduce phase then systematically aggregates these rationales across sibling nodes, resolving conflicts and reaching consensus at parent nodes. To more intuitively illustrate the MapReduce reasoning process, a case is provided in Figure~\ref{fig:case_reasoning_appendix}. 

\begin{figure*}[!t]
    \centering
    \includegraphics[width=\textwidth, height=\textheight, keepaspectratio]{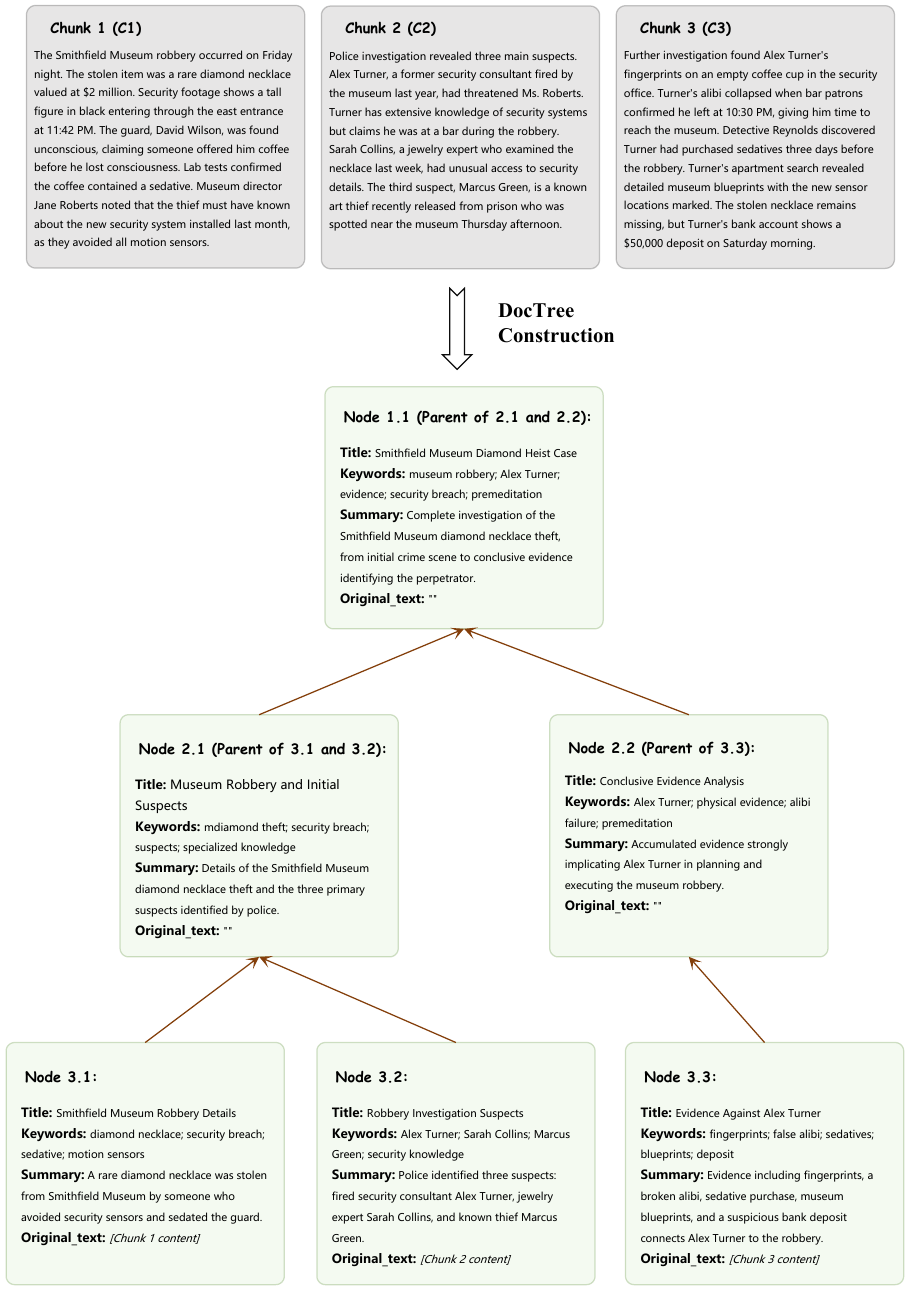}
    \caption{Case for DocTree Construction.}
    \label{fig:case_doctree_appendix}
\end{figure*}

This recursive process enables \method~to progressively refine understanding from leaf nodes to root nodes, preserving context coherence throughout the document hierarchy. By facilitating information flow between parent-child relationships and across sibling nodes, \method~transforms high-recall retrieved content into high-precision answers, particularly excelling at complex queries requiring integration of distributed information and resolution of apparent contradictions.

\subsection{Core Operational Prompts}
\label{Prompt_and_case}
\begin{figure*}[!t]
    \centering
    \includegraphics[width=0.8\linewidth]{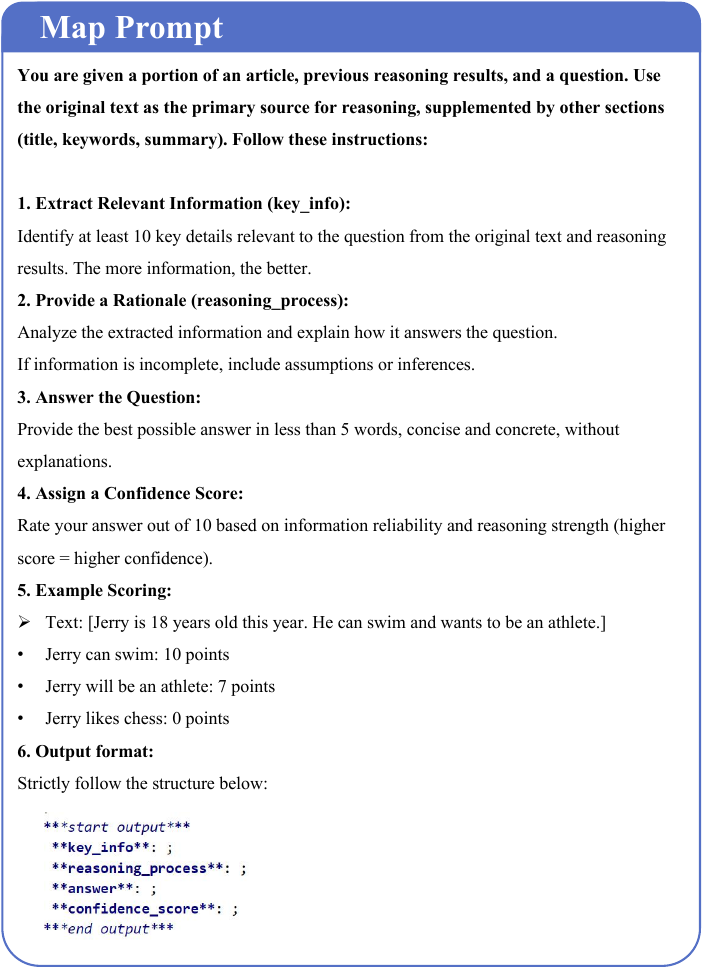}
    \caption{Prompt for Map step.}
    \label{fig:map_prompt_appendix}
\end{figure*}

\begin{figure*}[!t]
    \centering
    \includegraphics[width=0.8\linewidth]{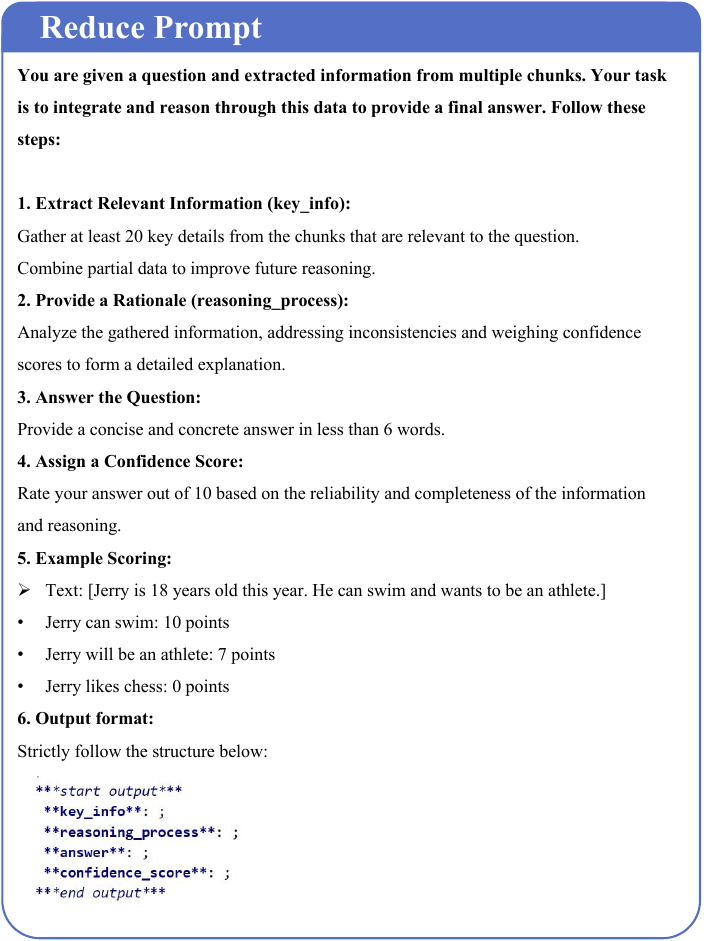}
    \caption{Prompt for Reduce step.}
    \label{fig:reduce_prompt_appendix}
\end{figure*}

\begin{figure*}[!t]
    \centering
    \includegraphics[width=0.8\linewidth]{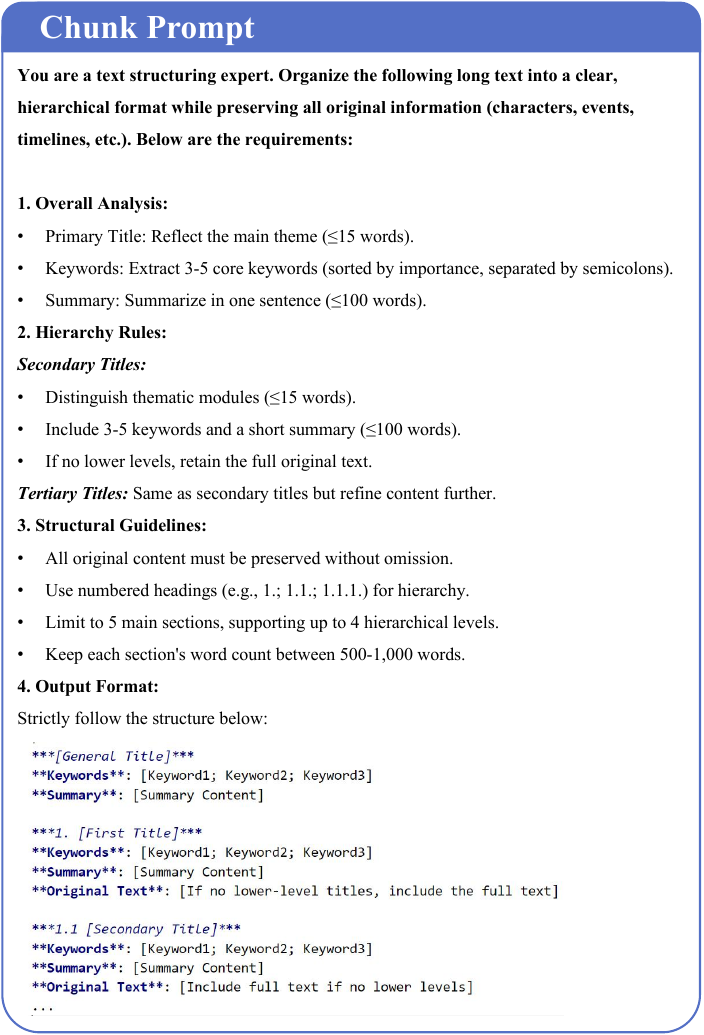}
    \caption{Prompt for Hierarchical Semantic Parsing.}
    \label{fig:chunk_prompt_appendix}
\end{figure*}

The Map, Reduce, and Hierarchical Semantic Parsing (HSP) steps in \method~are guided by specific prompts provided to the LLM. These prompts are crucial for structuring the LLM's processing and output at each stage.
Figure~\ref{fig:map_prompt_appendix} shows the prompt used for the Map step, Figure~\ref{fig:reduce_prompt_appendix} for the Reduce step, and Figure~\ref{fig:chunk_prompt_appendix} for the Hierarchical Semantic Parsing of chunks.

\subsection{Illustrative Case Studies}
\label{case_study}
To provide a concrete understanding of \method's internal workings, we present two case studies. Figure~\ref{fig:case_doctree_appendix} illustrates an example of DocTree construction. Figure~\ref{fig:case_reasoning_appendix} demonstrates the MapReduce reasoning process on a constructed DocTree. These examples visually walk through the key stages of our framework.

\end{document}